\documentclass[10pt,a4paper,twoside]{article}

\usepackage{latexsym,amssymb,amsmath,float}
\usepackage{graphicx}
\usepackage{multirow}
\usepackage{pifont}
\usepackage{array, adjustbox}

\newcolumntype{L}{>{\centering\arraybackslash}m{3cm}}

\newcommand{\cmark}{\ding{51}}
\newcommand{\xmark}{\ding{55}}

\begin{document}
\begin{center}{\Large\bf
A Self-Supervised Method for Body Part Segmentation and Keypoint Detection of Rat Images
}\end{center}
\begin{center}{\large\bf\noindent
László Kopácsi, Áron Fóthi, András Lőrincz
}\\[2mm]
Department of Artificial Intelligence, Faculty of Informatics, Eötvös Loránd University, Budapest, Hungary
\\[1mm]\texttt{
\{kopacsi,fa2,lorincz\}@inf.elte.hu
}\end{center}
\vspace*{7mm}

\begin{abstract}
Recognition of individual components and keypoint detection supported by instance segmentation is crucial to analyze the behavior of agents on the scene. Such systems could be used for surveillance, self-driving cars, and also for medical research, where behavior analysis of laboratory animals is used to confirm the aftereffects of a given medicine. A method capable of solving the aforementioned tasks usually requires a large amount of high-quality hand-annotated data, which takes time and money to produce. In this paper, we propose a method that alleviates the need for manual labeling of laboratory rats. To do so, first, we generate initial annotations with a computer vision-based approach, then through extensive augmentation, we train a deep neural network on the generated data. The final system is capable of instance segmentation, keypoint detection, and body part segmentation even when the objects are heavily occluded.
\end{abstract}

\section{Introduction}

Body part segmentation, keypoint detection, and instance segmentation are critical for understanding interactions between agents on the scene. To address these tasks, we have to tackle the difficulties arising from heavy occlusions of objects. Moreover, in the case of behavior prediction of laboratory animals, such as rats, objects can be highly similar as well.

There are several deep neural network-based architectures \cite{detr, hrnet, maskrcnn} capable of solving these tasks and handling the challenges posed by the dataset. However, they all require a large set of carefully annotated samples. Although there are solutions that can reduce the need for massive databases and make the annotation process faster \cite{deeplabcut, vv}, they still require some form of manual labeling. 

In this paper, we propose a method that is able to automatically annotate rats without any human effort. To realize this, an initial segmentation of objects is generated via foreground-background segmentation, from which keypoints and parts of samples are separated using various computer vision (CV) techniques. Given the derived data and their corresponding CV generated labels, deep neural networks are exploited for tracking. We use the Mask R-CNN \cite{maskrcnn} architecture both for body part segmentation and for keypoint detection (including the instance segmentation) tasks. We study different augmentation techniques and handle occlusions as suggested in the video segmentation literature \cite{lucid, masktrack}.

We evaluate the final method on hand-annotated samples using metrics introduced in the COCO benchmark \cite{coco}. We started with an initial average precision (AP) of 53.22\% on instance segmentation, 48.91\% on keypoint detection, and 9.38\% on body part segmentation, and by training deep models in a self-supervised manner, we achieved 61.92\%, 77.53\%, and 28.87\%, respectively.

Our contributions are listed below:
\begin{itemize}
    \item We propose a computer vision-based method for automatically annotating keypoints and body parts from foreground-background segmentation.
    \item We study the contributions of various augmentation methods used in the video segmentation literature.
    \item Finally, we train deep models on the generated labels in a self-supervised manner to handle the heavy occlusions present in the database.
\end{itemize}

\begin{figure}[h]
    \centering 
    \includegraphics[width=0.8\textwidth]{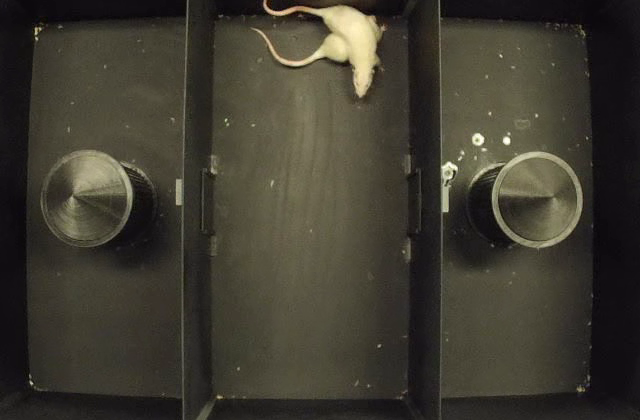}
    \caption{An example, where one of the rats is mounting the other one. The rats are highly similar and may heavily occlude each other from time to time giving rise to ambiguity in the instance segmentation and keypoint detection tasks due to hidden or heavily occluded regions.}
    \label{fig:challenges}
\end{figure}

\section{Proposed methods}
In this section, we first describe our proposed method for automatic annotation. Then we present the augmentation techniques and the architecture used for body part segmentation and keypoint detection. 

\begin{figure}[h]
    \centering 
    \includegraphics[width=\textwidth]{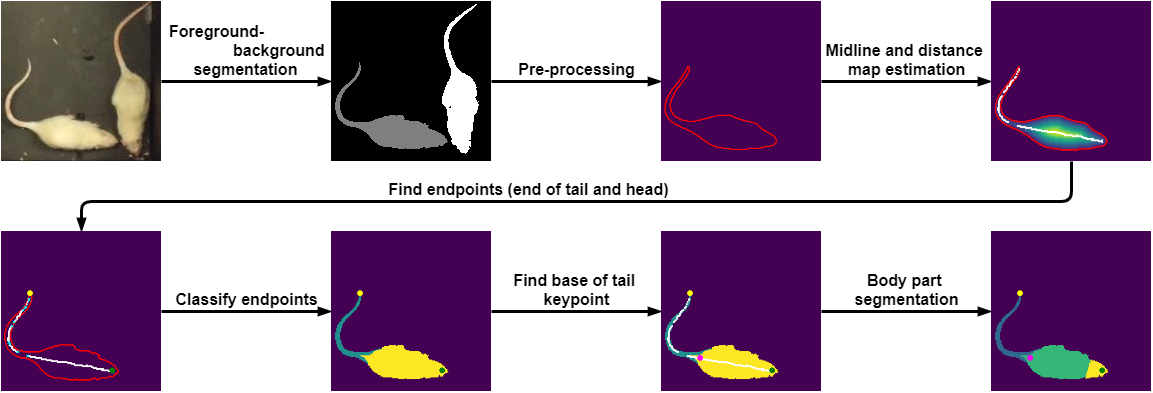}
    \caption{Pipeline of the automatic annotation method. The process starts with foreground-background segmentation; then, after pre-processing, medial axis transform is applied. From the endpoints of the midline, head and end of tail keypoints can be determined based on the area of their watershed segmentation. Then the base of the tail keypoint can be found by taking the median value of the distance transform. Finally, the body part segmentation is given by the watershed algorithm initiated from the acquired keypoints. Best viewed in color.}
    \label{fig:pipeline}
\end{figure}

The annotation process starts with the foreground-background segmentation (Section \ref{section:fgbg}), then we annotate each foreground segment by combining several computer vision methods (Section \ref{section:autoannot}). Figure \ref{fig:pipeline} shows the pipeline of the system.

By augmenting the generated samples (Section \ref{section:augment}) it is possible to train a deep model (Section \ref{section:architecture}) in a self-supervised manner.

\subsection{Foreground-background segmentation} \label{section:fgbg}
As the camera is stationary, it is possible to separate the foreground from the background by calculating the mode value of each pixel for multiple time steps then subtracting this value from the actual image.

Given a set of images $\mathcal{I} = \left\{ x_1, x_2, \dots, x_N \right\}$ recorded by a stationary camera, where $N$ is the number of images, and $x_i = [0, 255]^{H \times W \times 3}\ (i = 1, \dots, N)$ is a colored (RGB) image. The background can be estimated by 
\[B = \text{mode}\left(\begin{bmatrix}x_1 \\ x_2 \\ \dots \\ x_N\end{bmatrix},\ \text{axis}=0\right).\] 
Then the foreground of image $i$ can be determined by
\[F_i = x_i - B \qquad (i = 1, \dots, N).\]

\subsection{Automatic annotation} \label{section:autoannot}
For automatically annotating the foreground segments with keypoints and body parts, we combined several computer vision methods. We distinguished three keypoints: (i) the head, (ii) the base of the tail, and (iii) the end of the tail; and three body parts: (i) the head, (ii) the body, and (iii) the tail. The pipeline of the overall algorithm consists of 6 steps (Figure \ref{fig:pipeline}):

\begin{enumerate}
    \item Pre-processing: Improve initial segmentation and decide whether it is possible to annotate the mask.
    \item Midline and distance map estimation: Estimate them by using medial axis transformation \cite{zhang1984skeleton}. 
    \item Find endpoints (end of tail and head): Find a pair of connected points via the longest minimum cost path given the midline.
    \item Classify endpoints: By applying the watershed algorithm \cite{watershed}, the endpoints can be classified, given the area of the resulting segments.
    \item Find base of tail keypoint: The last keypoint can be estimated by the location of the distance map's median value on the midline.
    \item Body part segmentation: The body parts of the rat can be determined by using another watershed algorithm.
\end{enumerate}

\paragraph{Pre-processing:} 
The resulting masks of the foreground-background segmentation can be noisy. In order to use them, we need to do some pre-processing. To this end, we need to 
\begin{enumerate}
    \item Clean the mask: first we remove holes, then apply closing and finally remove torn off segments, then
    \item Decide whether it is possible to annotate the mask: this can be checked by measuring its convexity, and finally
    \item Smooth the boundary: otherwise, the resulting skeleton from the medial axis transform may contain spurious line segments and spoil our midline estimation. Therefore, we smooth the boundary by fitting 4th degree B-splines to it.  
\end{enumerate}

\paragraph{Midline and distance map estimation:} 
After pre-processing, the skeleton and the distance transform of the mask are determined by the medial axis transform \cite{zhang1984skeleton}. The distance transform is a grayscale image, where the intensity values represent the distance from the boundary, while the skeleton is a binary image, a remnant of the original mask. For each skeleton point, the corresponding distance transform value represents the radius of the largest circle, which is centered at that point and touches at least two boundary points. All other points of the circle are within or at the boundary of the mask. As this process is highly sensitive to irregularities in the boundary, proper smoothing is required to give us an accurate midline estimate.

We used the implementation of the medial axis transformation from the scikit-image Python package.

\paragraph{Find endpoints (end of tail and head):}
From the endpoints of the skeleton, we need to find those two, which are the furthest from each other. As the animals can take arbitrary poses, a simple pixel distance will not give optimal results. The endpoints should be selected by finding the longest minimum-cost path between them, given the midline.

Let $\mathcal{P} = \left\{ p_1, p_2, \dots, p_K \right\}$ a set of all endpoints of the medial axis transform and $D = \mathbb{N}^{H \times W}$ the cost matrix. Each endpoint represents a coordinate of the image: $p_k = (h, w)$, where $h \in [1, H] \land w \in [1, W]$, and all endpoints have a cost of one:
\[ D_{i,j} = 
    \begin{cases}
        1, & \text{if}\ (i, j) \in \mathcal{P} \\
        \epsilon, & \text{otherwise}
    \end{cases}, \]
where $1 < \epsilon \in \mathbb{N}$.
In this case, the end of the tail and the head keypoints can be found using the following formula
\[ (q, r) = \mathop{\mathrm{argmax}}_{p_i, p_j \in \mathcal{P} \land p_i \neq p_j} \text{MCD}(p_i, p_j, D), \]
where $\text{MCD}(., ., .)$ calculates the minimum-cost path between the specified coordinates given the cost matrix. 

\paragraph{Classify endpoints:}
To determine which endpoint corresponds to the head and which one to the end of the tail, we apply the watershed \cite{watershed} algorithm using the negative distance transform and initializing the basins with the two endpoints. Then we classify the point with the larger segmented area as the head and the other one as the end of the tail keypoint.

We used the implementation of the watershed algorithm from the scikit-image Python package.

\paragraph{Find base of tail keypoint:}
After we have acquired the head and the tail keypoints, the final keypoint location is determined by the position of the distance transform's median value on the midline that separates  the larger and the smaller values.

We also perform an additional validation step in order to exclude conceivably incorrect annotations. By measuring the tail-to-body ratio \cite{bodytailratio}, we can decide whether the tail is partially missing due to foreground-background segmentation mistakes or not and dismiss the mask if necessary.

\paragraph{Body part segmentation:}
Given all the keypoints, we can segment the body parts by applying another watershed algorithm. 

\begin{figure}[h]
    \centering 
    \includegraphics[width=0.8\textwidth]{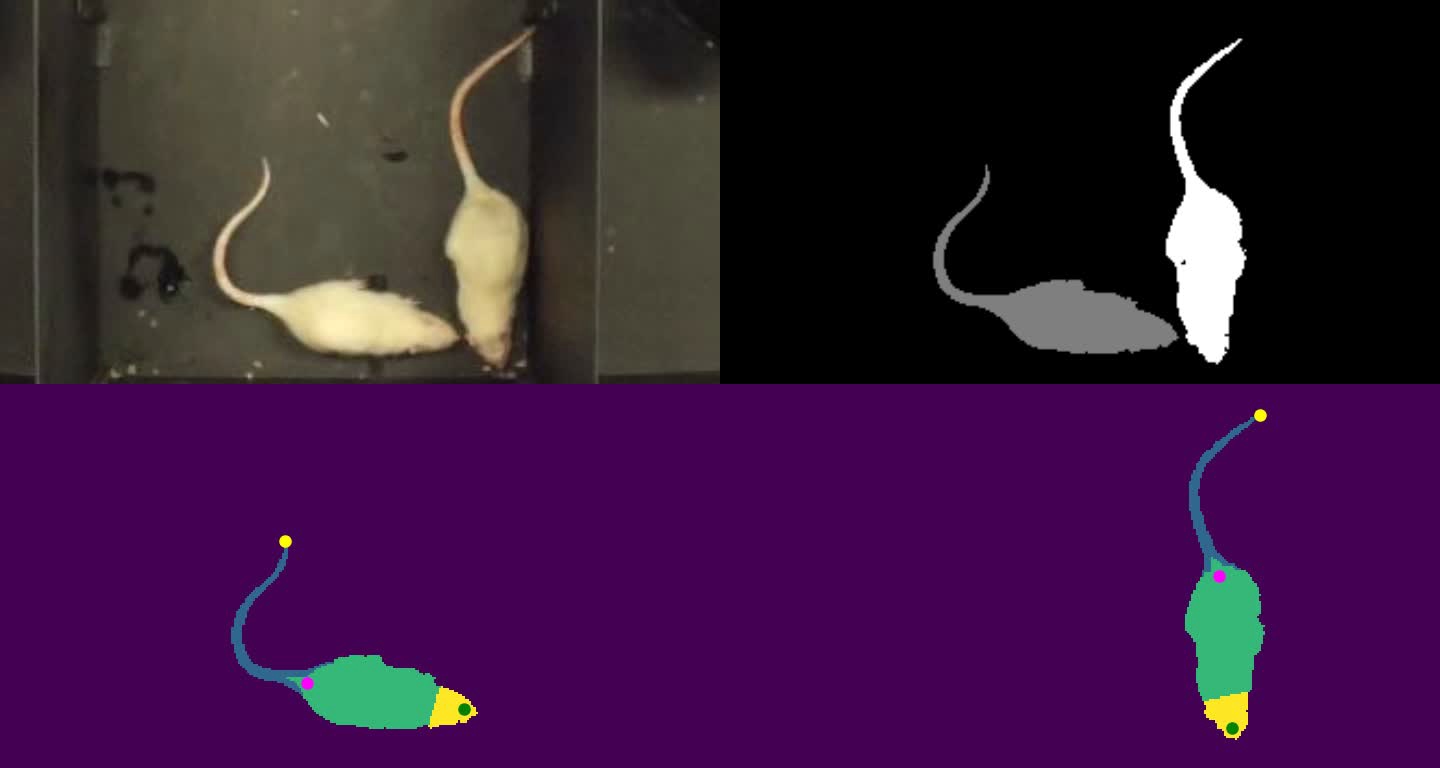}
    \caption{Result of the CV-based automatic annotation method. In the first row the input image and its foreground-background segmentation, and in the second row the annotated instances can be seen.}
    \label{fig:autoannot}
\end{figure}

Finally, we can do some post-processing steps to improve the results by placing the keypoints close to the nearest corner of the mask and trimming protruding parts of the head segment. Figure \ref{fig:autoannot} shows the final result of the automatic annotation pipeline.

\subsection{Augmentation} \label{section:augment}
Since the proposed computer vision-based automatic annotation method (Section \ref{section:autoannot}) is only capable of annotating rats if they are not occluded, we need to introduce occlusions in the dataset by augmenting the generated training samples. Otherwise, the trained model could not exceed the performance of the initial method. To simulate occlusions, we randomly cut out one of the rats, fill its place in with the background, then shift the segment near the other instance. To achieve the best results, we combined several augmentation techniques commonly used in the video segmentation literature \cite{lucid, masktrack}. We
\begin{itemize}
    \item applied various kind of smoothing: Gaussian, median pyramid as well as  no smoothing,
    \item rotated objects around its center by $-45^\circ$ to $45^\circ$,
    \item scaled the shifted segment by $0.9$ to $1.1$, and
    \item deformed the mask via thin-plate-spline warping \cite{warping}.
\end{itemize}

\subsection{Architecture overview} \label{section:architecture}
\begin{figure}[h]
    \centering 
    \includegraphics[width=0.6\textwidth]{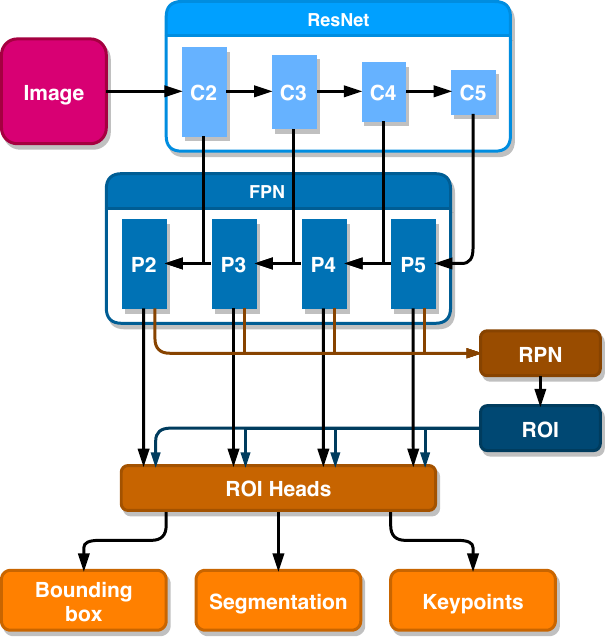}
    \caption{Mask R-CNN architecture. The network takes an RGB image, feeds it through the backbone, which extracts a feature pyramid, where each feature map has half the spatial resolution as the one before. Then the RPN makes bounding box proposals, which are processed by the ROI heads.}
    \label{fig:maskrcnn}
\end{figure}

Several deep architectures are capable of instance segmentation and keypoint detection, such as \cite{detr, hrnet, maskrcnn}.
For ease of use and modularity, we chose the Mask R-CNN architecture (see Figure \ref{fig:maskrcnn}). This model consists of 3 main parts:
\begin{itemize}
    \item The backbone network is typically a ResNet50 \cite{resnet} wrapped in a feature pyramid network (FPN) \cite{fpn}. It accepts an RGB image as its input and extracts a set of feature maps with different spatial resolutions.
    \item The region proposal network (RPN) is inputted by the feature maps and generates bounding box proposals based on their objectness score, then
    \item The region of interest (ROI) heads process the proposed boxes and return the final results. 
\end{itemize}
By combining different types of ROI heads, we can produce a model that is capable of instance segmentation and keypoint detection, and another one which is able to detect and segment body parts.

\section{Results}
Here, we introduce the dataset and the performance metric and present the results of our proposed methods.

\subsection{Experimental setup}
We implemented the automatic annotation method in Python using the scikit-image image processing library and NumPy. For the deep learning model, we used the Mask R-CNN implementation of Detectron2\footnote{https://github.com/facebookresearch/detectron2}.

We used ResNet50 with FPN for the backbone, like in the original Mask R-CNN architecture \cite{maskrcnn}. We initialized the weights of our model with the COCO instance segmentation baseline in the case of body part segmentation, and with the COCO person keypoint detection baseline in the case of instance segmentation and keypoint detection. During training, we used a batch size of 512 for the heads and a batch size of 4 for the rest of the network. We initialized the learning rate to 0.00025 and set the object keypoint similarity (OKS) sigma values to 0.079, 0.107, and 0.089 for the head, base of the tail, and end of the tail keypoints, respectively. These sigma values correspond to the shoulders, hips, and ankles in the original COCO benchmark \cite{coco}. We trained each model for a maximum of 50,000 iterations on an NVIDIA GeForce GTX 1080 graphics card. During the evaluation, we set the confidence threshold to 70\%. All other parameters were set to the default values.  

Our implementation is available on GitHub\footnote{https://github.com/lkopi/rat\_segmentation}.

\subsection{Database}
The database contains a continuous video of 2 white rats in a black container. The video was recorded by a stationary camera fixed on the top of the container. The video is 58994 frame-long, and the original resolution was 1280x720, from which we cropped a 640x420 region containing only the box. We only used the first 15,000 frames for training, and we annotated 200 frames, where no occlusion was present for the sake of evaluation.

The main challenges posed by the dataset are (i) the lack of ground truth labels, (ii) the highly similar objects, and (iii) the large amount of occlusion. (Figure \ref{fig:challenges}.) To attain good results, one should address all of these problems.

\paragraph{Benchmark measure}
To measure the performance of our method, we use the metrics of the COCO Detection and Keypoint benchmark \cite{coco}. We evaluated the methods on the hand-annotated samples. We use average precision (AP) as our primary performance measure. AP is calculated by averaging over 10 Intersection over Union (IoU) levels from 0.5 to 0.95 with a step size of 0.05. The value of the AP is ranging between 0 and 1, and the higher, the better. It can also be calculated for keypoints, but in this case, the Object Keypoint Similarity (OKS) is used instead of IoU.

\subsection{Automatic annotation}
The proposed automatic annotation method is composed of several CV methods. Most of them can be tuned by setting their parameters or being replaced by alternative approaches, such as using active contours \cite{snakes} instead of B-splines to smooth object boundary. We tried several combinations of parameters and methods from which a few are presented in Table \ref{tbl:autoannot}. 

We got the best overall result when we were using the method described in Section \ref{section:autoannot}. On keypoint detection and instance segmentation, it reached 48.91 and 53.22 AP, respectively, while on body part segmentation, the AP was merely 9.38. Nevertheless, with proper augmentation, it proved to be enough for training deep neural networks. In the rest of the paper, we refer to this as baseline or CV-based approach. Figure \ref{fig:autoannot} shows the result of this method.

\begin{table}[H]
    \centering
    \begin{adjustbox}{width=1.0\textwidth, center}
        \begin{tabular}{|ccc||ccc||cc|}
            \hline
            \multicolumn{1}{|m{1.8cm}}{\centering\textbf{Set head to nearest corner}} &
                \multicolumn{1}{m{1.3cm}}{\centering\textbf{Clean mask}} &
                \multicolumn{1}{m{2cm}||}{\centering\textbf{Trim protruding parts}} &
                \multicolumn{1}{m{1.8cm}}{\centering\textbf{Bounding box AP}} &
                \multicolumn{1}{m{1.8cm}}{\centering\textbf{Keypoint AP}} &
                \multicolumn{1}{m{2.4cm}||}{\centering\textbf{Segmentation AP}} &
                \multicolumn{1}{m{1.8cm}}{\centering\textbf{Bounding box AP}} &
                \multicolumn{1}{m{2.4cm}|}{\centering\textbf{Segmentation AP}} \\
            \hline
            \cmark & \xmark & \cmark & 56.76 & 47.36 & 44.00 & 11.81 & 9.37 \\
            \xmark & \xmark & \cmark & 68.13 & 47.76 & 53.03 & 16.59 & 9.37 \\
            \xmark & \cmark & \cmark & \textbf{67.71} & \textbf{48.91} & \textbf{53.22} & \textbf{17.40} & \textbf{9.38} \\
            \xmark & \cmark & \xmark & 67.90 & 48.83 & 53.42 & 16.20 & 9.38 \\
            \hline
        \end{tabular}
    \end{adjustbox}
    \caption{Results of the automatic annotation method. The first three columns show some of the system's parameters. The subsequent three and the last two columns show the performance on the keypoint detection and instance segmentation task, and on the body part segmentation task. We achieved the best overall performance with the method highlighted in bold. The details are described in Section \ref{section:autoannot}.}
    \label{tbl:autoannot}
\end{table}

\subsection{Self-supervised method}
We trained two separate models on the data produced by the automatic annotation method. The task of the first one was keypoint detection and instance segmentation, while the second model was trained to detect and segment body parts. The hand-annotated labels were not introduced in the training process; they were only used to evaluate the performance of the final models. On both tasks, the trained models outperformed the CV-based approach. On keypoint detection, it was better by 35.25\%, while on body part segmentation, it surpassed the baseline by 2.35 times, contributing to a \~135\% improvement.

\begin{figure}[H]
    \centering 
    \includegraphics[width=\textwidth]{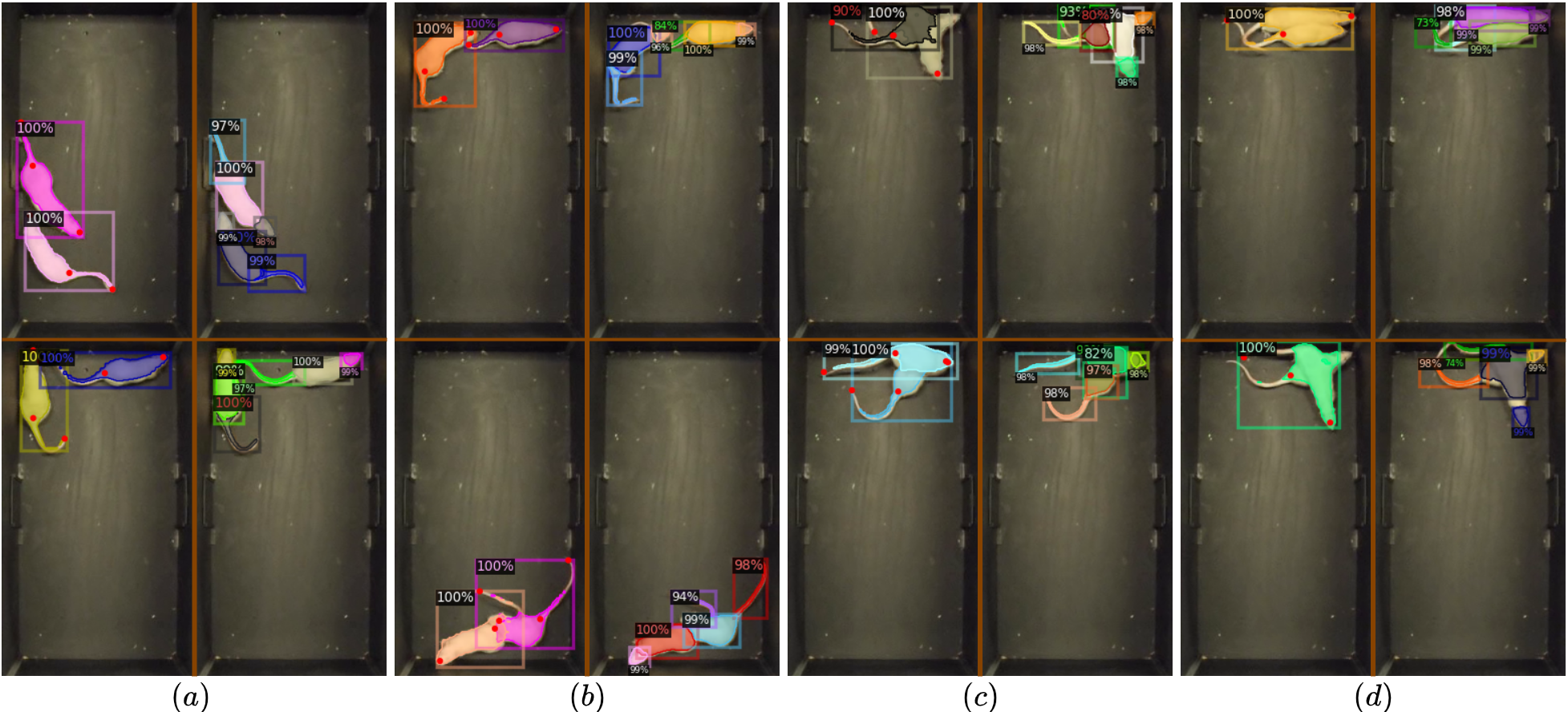}
    \caption{Result of the final method. The left side of each image shows the result of the keypoint detection and instance segmentation model, and the right side shows the result of the body part segmentation model. The columns address the different amount of occlusion present on the scene. If there is no occlusion $(a)$ or just partial occlusion $(b)$ is present on the scene, both methods perform well. If the objects are heavily occluded $(c)$, the quality of the keypoint model starts degrading. After a point, $(d)$, it cannot separate instances due to limitations of the Mask R-CNN architecture, but the body part model can still segment the objects.}
    \label{fig:results}
\end{figure}

Figure \ref{fig:results} shows some outputs of the final models. Both models perform well even in annotating heavily occluded objects, but the keypoint detection model cannot detect objects when they are very close to each other. This issue is due to the architecture because Mask R-CNN uses non-maximum suppression to choose the best fitting bounding box for each object. Nevertheless, in such cases, the body part model can manage the segmentation of the object.    

\subsubsection{Keypoint detection and instance segmentation model}
After 30K iterations, the performance of the networks, which used pyramid median or Gaussian smoothing, stagnated. While without smoothing, we were able to train the model for 50K iterations. In all cases, we observed that by applying more augmentation, better results could be reached. We got the best outcomes when no smoothing was applied. However, these models needed much more iterations to compete with others. By trained only for 30K iterations, their performance was worse than using just basic augmentation. It reached only 82.57, 47.79, and 60.09 AP. However, after training for 20K more iterations, it outperformed all previous models. The final results are 90.27, 61.92, and 77.53 AP on object detection, keypoint detection, and instance segmentation, respectively. Compared to the baseline (CV method), it achieved a 35.25\% improvement. We summarized the results in Table \ref{tbl:kp}. 

\begin{table}
    \centering
    \begin{adjustbox}{width=0.9\textwidth, center}
        \begin{tabular}{|cccc||ccc|}
            \hline
            \multicolumn{1}{|m{2cm}}{\centering\textbf{Rotation and scaling}} &
                \multicolumn{1}{m{1.5cm}}{\centering\textbf{Warping}} &
                \multicolumn{1}{m{1.8cm}}{\centering\textbf{Smoothing}} &
                \multicolumn{1}{m{1.8cm}||}{\centering\textbf{\# of iterations}} &
                \multicolumn{1}{m{1.8cm}}{\centering\textbf{Bounding box AP}} &
                \multicolumn{1}{m{1.8cm}}{\centering\textbf{Keypoint AP}} &
                \multicolumn{1}{m{2.4cm}|}{\centering\textbf{Segmentation AP}} \\
            \hline
            \xmark & \xmark & None & 30K & 84.04 & 58.75 & 60.77 \\
            \cmark & \xmark & None & 50K & 85.30 & 61.30 & 63.46 \\
            \cmark & \cmark & None & 50K & \textbf{90.27} & \textbf{61.92} & \textbf{77.53} \\
            \hline
            \xmark & \xmark & Pyramid & 30K & 83.98 & 53.01 & 60.56 \\
            \cmark & \xmark & Pyramid & 30K & 85.58 & 51.47 & 59.01 \\
            \cmark & \cmark & Pyramid & 30K & 83.43 & 56.13 & 59.45 \\
            \hline
            \xmark & \xmark & Gauss & 30K & 83.17 & 43.13 & 59.96 \\
            \cmark & \xmark & Gauss & 30K & 84.26 & 55.99 & 60.12 \\
            \cmark & \cmark & Gauss & 30K & 85.30 & 61.73 & 60.58 \\
            \hline
        \end{tabular}
    \end{adjustbox}
    \caption{Results of the trained model on the keypoint detection and instance segmentation task. We studied the effect of different augmentation techniques presented in Section \ref{section:augment}. Warping denotes thin-plate-spline warping with a keypoint shift of a maximum of 10 pixels. The best performance was acquired when no smoothing was applied. In all cases, the more augmentation we used, the better the result got.}
    \label{tbl:kp}
\end{table}

\subsubsection{Body part segmentation model}
In the case of body part segmentation (Table \ref{tbl:bp}), we observed the clues of over-fitting in a much earlier state. Models trained only for 5K iterations outperformed all subsequent stages. We discovered similar insights as in the keypoint detection case. In general, models trained with tedious augmentation achieved similar or slightly better results except in the no smoothing experiments, where the performance decreased when thin-plate-spline warping was applied. 
The added value of augmentation was modest ($\pm$(0-6)\%) as the applied techniques do not affect parts significantly. Consequently, more augmentation might be needed to reach better results.

We reached the best performance when only rotation and scaling were used without any smoothing. The model achieved 34.01 AP on body part detection and 28.87 AP on segmentation, which accounts for a 134.78\% gain compared to the CV-based approach.

\begin{table}[h]
    \centering
    \begin{adjustbox}{width=0.8\textwidth, center}
        \begin{tabular}{|cccc||cc|}
            \hline
            \multicolumn{1}{|m{2cm}}{\centering\textbf{Rotation and scaling}} &
                \multicolumn{1}{m{1.5cm}}{\centering\textbf{Warping}} &
                \multicolumn{1}{m{1.8cm}}{\centering\textbf{Smoothing}} &
                \multicolumn{1}{m{1.8cm}||}{\centering\textbf{\# of iterations}} &
                \multicolumn{1}{m{1.8cm}}{\centering\textbf{Bounding box AP}} &
                \multicolumn{1}{m{2.4cm}|}{\centering\textbf{Segmentation AP}} \\
            \hline
            \xmark & \xmark & None & 5K & 30.50 & 28.77 \\
            \cmark & \xmark & None & 5K & \textbf{34.01} & \textbf{28.87} \\
            \cmark & \cmark & None & 5K & 32.27 & 29.17 \\
            \hline
            \xmark & \xmark & Pyramid & 5K & 32.60 & 28.98 \\
            \cmark & \xmark & Pyramid & 5K & 32.32 & 27.87 \\
            \cmark & \cmark & Pyramid & 5K & 33.07 & 29.10 \\
            \hline
            \xmark & \xmark & Gauss & 5K & 32.53 & 27.93 \\
            \cmark & \xmark & Gauss & 5K & 32.23 & 27.88 \\
            \cmark & \cmark & Gauss & 5K & 33.13 & 27.51 \\
            \hline
        \end{tabular}
    \end{adjustbox}
    \caption{Results on the body part segmentation tasks. We applied similar augmentation techniques to the keypoint detection task. As the results show, their contribution was smaller as they did not significantly affect individual components. During training, the performance of the models dwindled after 5,000 iterations, so more rigorous augmentation might be necessary to further improve the AP. The best overall result was achieved when only rotation and scaling was used. This method is marked in bold typefaces.}
    \label{tbl:bp}
\end{table}

\section{Summary}
In this paper, we presented a method capable of instance segmentation, keypoint detection, and body part segmentation without the need for any hand-annotated data. We studied the effect of different augmentation techniques and achieved AP of 61.92\%, 77.53\%, and 28.87\% on instance segmentation, keypoint detection, and body part segmentation, respectively.

Our results show that automatic annotation of rat images is possible without the need for manual labeling. The method presented here can be used to analyze the interaction between instances and to speed up the annotation process if more precise labels are needed. To improve the performance, one should address the shortcoming of Mask R-CNN in the case of instance segmentation, which is due to non-maximum suppression, by changing the architecture \cite{detr} or by separating bounding boxes if more than one instance is present. 

The system could be extended to handle video sequences via bipartite matching \cite{tracktor} or by incorporating optical flow estimated by deep neural networks \cite{pwcnet}, among others. In addition, the power of such temporal extension methods and the fusion of the keypoint and the body part models may be able to track such very similar and flexible objects subject to heavy occlusions in order to study their interactions in an automated fashion.

\subsection*{Acknowledgements}
The research has been supported by the European Institute of Innovation and Technology. We want to thank Árpád Dobolyi and Dávid Keller for providing the database. Á.F. and A.L. were supported by the ELTE Institutional Excellence Program of the National Research, Development and Innovation Office (NKFIH-1157-8/2019-DT), and by the ``Application Domain Specific Highly Reliable IT Solutions'' project implemented with the support of the National Research, Development and Innovation Fund of Hungary under the Thematic Excellence Programme no. 2020-4.1.1.-TKP2020 (National Challenges Subprogramme) funding scheme, respectively. 

\subsection*{Author Contributions}
L.K. and A.L. conceived and designed the research, L.K. and Á.F. performed computational analyses.

\end{document}